\documentclass[10pt,twocolumn,letterpaper]{article}

\usepackage[pagenumbers]{wacv} % To force page numbers, e.g. for an arXiv version

\usepackage{graphicx}
\usepackage{amsmath}
\usepackage{amssymb}
\usepackage{booktabs}
\usepackage{placeins}
\usepackage{gensymb}
\usepackage[dvipsnames]{xcolor}

\newcommand{\redtext}[1]{\textcolor{red}{#1}}
\newcommand{\bluetext}[1]{\textcolor{blue}{#1}}
\newcommand{\greentext}[1]{\textcolor{green}{#1}}
\newcommand{\pinktext}[1]{\textcolor{Lavender}{#1}}
\usepackage[pagebackref,breaklinks,colorlinks]{hyperref}

\usepackage[capitalize]{cleveref}
\crefname{section}{Sec.}{Secs.}
\Crefname{section}{Section}{Sections}
\Crefname{table}{Table}{Tables}
\crefname{table}{Tab.}{Tabs.}

\begin{document}

\title{Long-Term 3D Point Tracking By Cost Volume Fusion}

\author{Hung Nguyen
\and 
Chanho Kim
\and
Rigved Naukarkar
\and 
Li Fuxin
\and 
Oregon State University\\
{\tt\small \{nguyehu5, kimchanh, naukarkr, lif\}@oregonstate.edu}
}
\maketitle

%%%%%%%%% ABSTRACT
\begin{abstract}
Long-term point tracking is essential to understand  non-rigid motion in the physical world better. Deep learning approaches have recently been incorporated into long-term point tracking, but most prior work predominantly functions in 2D. Although these methods benefit from the well-established backbones and matching frameworks, the motions they produce do not always make sense in the 3D physical world. In this paper, we propose the first deep learning framework for long-term point tracking in 3D that generalizes to new points and videos without requiring test-time fine-tuning. Our model contains a cost volume fusion module that effectively integrates multiple past appearances and motion information via a transformer architecture, significantly enhancing overall tracking performance. In terms of 3D tracking performance, our model significantly outperforms simple scene flow chaining and previous 2D point tracking methods, even if one uses ground truth depth and camera pose to backproject 2D point tracks in a synthetic scenario.
  % \keywords{3D Point Tracking \and Point cloud \and Scene Flow \and Motion Estimation }
\end{abstract}

%%%%%%%%% BODY TEXT
\section{Introduction}
\label{sec:intro}
Motion estimation is a task that has existed since the beginning of computer vision. Short-term dense motion estimation problems,
such as optical flow in 2D and scene flow in 3D, have been extensively studied. 
However, the utility of these tasks is limited because  many points are featureless, making their motion estimation within two frames fundamentally ambiguous.
Besides,  chaining 2-frame motions to derive point trajectories is susceptible to significant cumulative errors and ineffective for handling long occlusions.

However, spatial-temporal tracking of keypoints have always been interesting~\cite{laptev2005space} because it offers the capability to track long-term non-rigid object motions. Such knowledge would be greatly helpful for augmented reality and robotics applications, as well as providing supervision for generative models that generate dynamic videos with arbitrary non-rigid motions. 
Recently, the authors of~\cite{harley2022particle} reinvigorated the long-term pixel tracking problem and proposed a framework inspired by previous state-of-the-art optical flow and object tracking work. \cite{doersch2022tap} released datasets specifically designed to address point tracking. These datasets, including 2D and 3D data, have boosted research on this task. 

Most existing methods predominantly address the problem of  2D long-term point tracking. But in the 3D world we live in,  such 2D tracking, however accurate it might be, might still miss the 3D motion of the points. 
%This could lead to issues in downstream tasks such as the reconstruction of deformable 3D objects, robotic object manipulation and physics simulation based on those motion estimates. 
Even the state-of-the-art video generator SORA has significant trouble understanding long-term 3D point  motion~\cite{bupe2024wrong}, which may  limit its usage for downstream tasks such as augmented reality and robot manipulation. As we will show in the experiments, backprojecting 2D long-term point tracks into 3D, even with known camera poses and pixel depths, is still prone to errors due to the sparsity of the depth maps, occlusion, and the accumulation of numerical errors.

The ability to track any point long-term in 3D would significantly enhance  our understanding of the scene dynamics. %However, prior work on this has mostly only addressed the 2-frame case as in the scene flow problem. 
Recently, 
\cite{luiten2023dynamic} proposed a test-time optimization method to model dynamic scenes using a set of Gaussians, thereby enabling long-term point tracking. While surpassing the performance of prior 2D methods, this approach relies on test-time optimization for each scene and 
%
%has inherent drawbacks. Its test-time optimization nature mitigates domain gap issues but at the cost of prolonged processing times for each scene. Additionally, the framework, due to imposed constraints, 
struggles to track new points entering the scene. Test-time optimization is quite computationally expensive, especially for longer videos, making it unsuitable for online tracking.
%or novel scenes, posing challenges for real-life applications.

Recognizing the potential of tracking directly in 3D space, we propose a simple yet efficient method for long-term \textbf{online} tracking of keypoints in a dynamic 3D point cloud. To our knowledge our approach is the first deep learning framework that directly attacks long-term 3D point tracking in a generalizable manner that can work on new points and videos \textbf{without} test-time fine-tuning. 
Our online tracking framework takes as input a sequence of point clouds representing the dynamic scene. The model predicts a position for each query point by combining multiple past appearance information with motion information from the past point trajectory with a transformer-based framework. Occlusions are predicted explicitly to filter out noisy appearance features.
We propose an adaptive decoding module that selectively decodes around the query point, enabling the network to process denser point clouds and generate more precise motion for each point. Experiment results show that our approach significantly outperforms baselines such as linking scene flow results or 3D tracks backprojected from 2D point tracks obtained by prior work.

In summary, our contributions include:
\begin{itemize}
\vspace{-0.03in}
    \item We propose the first online deep learning-based tracking framework that can track any point in 3D  \textbf{without} test-time optimization.
    \vspace{-0.05in}
    \item We devise a novel Cost Volume Fusion module that effectively takes into account the long-term appearances of each point and its past motion trajectory.\vspace{-0.05in}
    \item We propose an adaptive decoding module that significantly reduces memory consumption when training on denser point clouds, allowing the model to produce better motion predictions.
    \vspace{-0.03in}
\end{itemize}

%------------------------------------------------------------------------

\section{Related Work}
\label{related_work}
% \redtext{Update the related work with 3 new papers in SLACK}
\subsection{Point tracking}
Tracking any pixel or point in 2D long-term has recently gained significant attention. MFT~\cite{schmidt2023tracking} presents an extension for optical flow by constructing optical flows not only between consecutive frames but also between distant frames. These flows are chained together guided by the predicted occlusion and uncertainty scores obtained from pre-trained networks, to derive the most reliable sequence of flows for each tracked pixel.
% \redtext{These 2 paragraphs can be combined and shorten if we need more space}
PIPs~\cite{harley2022particle} presents a novel framework designed for multi-frame point tracking 
% This framework employs a sliding-window approach and concurrently updates the target point's position in each frame, enabling effective tracking even during brief occlusion events. 
which simultaneously estimates the target point positions in multiple frames. Therefore, it can handle short occlusion events.
To improve tracking performance, Cotracker~\cite{karaev2023cotracker} utilizes self-attention layers to track target points and their local and global contextual points together. The self-attention layers enable information exchange among these points, leading to better tracking quality.

% the correlation among various points. The authors suggest the simultaneous tracking of the target point along with select local and global contextual points. The model incorporates several self-attention layers to exchange the motion information among the target and context points.

Tap-Vid~\cite{doersch2022tap} 
%redefines the problem of "tracking any point" and 
offers valuable datasets along with a straightforward baseline method. It achieves this by predicting the position of a point in each frame based on the cost volume derived from the query feature and the corresponding feature map. Meanwhile, TAPIR~\cite{doersch2023tapir} introduces a two-stage network comprising a matching stage and a refinement stage. This framework also incorporates the prediction of uncertainty to suppress ambiguous or unreliable predictions, enhancing point tracking accuracy.

% \redtext{Briefly Discuss Neural Non-Rigid Tracking}
\cite{bozic2020neural} introduces a differentiable non-rigid approach that achieves superior performance in reconstructing non-rigidly moving objects. However, this approach can only focus on a single object in the video. In contrast, we focus on modeling entire dynamic scenes by tracking any point within them.

A different line of work utilizes test-time optimization techniques to model the scene, resulting in superior tracking performance. Additionally, these methods directly track points in 3D. Therefore, they can utilize useful 3D priors for tracking purposes. 
%These methods usually trade-off between the running time for the scene modeling and the tracking performance.
% OmniMotion~\cite{wang2023omnimotion} introduces a test-time optimization method. 
Specifically, OmniMotion~\cite{wang2023omnimotion} represents video content by employing a canonical 3D volume. It learns a set of bijections to map points between any frame and the canonical one, thereby enabling tracking capability.
%Tracking is accomplished through the mapping between local and canonical volumes, facilitated by a set of learned bijections designed to establish correspondence between any point within a frame's local volume and the canonical volume, and vice versa. 
Similarly, \cite{luiten2023dynamic} represents the scene with a set of Gaussians. While the number of Gaussians, their colors, and opacity remain fixed throughout the video, these Gaussians can move and rotate freely to model the dynamic scene. Therefore, the tracking capability emerges from persistently modeling the dynamic scene under these constraints.

To achieve superior tracking performance, \cite{wang2023omnimotion, luiten2023dynamic} require significant time to reconstruct the 3D model of the scene using test-time optimization. Consequently, these methods cannot be used for online tracking. In contrast, our method tracks points online without test-time optimization. 

\subsection{Scene flow estimation}
Scene flow estimation predicts the 3D motion field of points, with the input being either 2D images or 3D point clouds. We focus on the approaches that has 3D point clouds as input as they are more relevant to our work. Approaches can usually be categorized into two types: supervised and self-supervised. However, many supervised methods can utilize self-supervised losses to adapt a pre-trained model to a new dataset without ground truth.

FlowNet3D~\cite{liu2019flownet3d} is among the first to use PointNet++~\cite{qi2017pointnet++}, a deep network for point clouds, to predict scene flow from point clouds directly and proposes important basic modules such as flow embedding, set conv, and set upconv layers that are commonly used in subsequent works. FLOT~\cite{puy2020flot} reformalizes scene flow estimation as an optimal transport problem to initially compute scene flow and subsequently refine it using a deep network. PointPWC~\cite{wu2020pointpwc} introduces novel cost volumes, upsampling, and warping layers and utilizes a point convolution network~\cite{wu2019pointconv} to handle 3D point cloud data. Scene flow is then constructed in a coarse-to-fine fashion. The authors also propose a novel self-supervised loss that has been adopted in subsequent works~\cite{zhang2024seflow,wang2021unsupervised, fang2024self, shen2023self}. PV-Raft~\cite{wei2021pv} introduces a point-voxel correlation field to handle both long-range and local interactions between point pairs. 
NSFP~\cite{li2021neural} represents scene flow implicitly with a neural network trained directly on test scenes with self-supervised losses. Fast neural scene flow~\cite{Li_2023_ICCV} improves upon NSFP by utilizing the distance transform~\cite{breu1995linear, danielsson1980euclidean} and a correspondence-free loss, significantly reducing processing time.
Finally, with the rise of diffusion models,~\cite{zhang2024diffsf, liu2024difflow3d} incorporate diffusion processes into the scene flow estimation pipeline, achieving millimeter-level end-point error.
%------------------------------------------------------------------------

\section{Method}
\subsection{Overview}
\begin{figure*}[t]
  \centering
   \includegraphics[width=1.\linewidth]{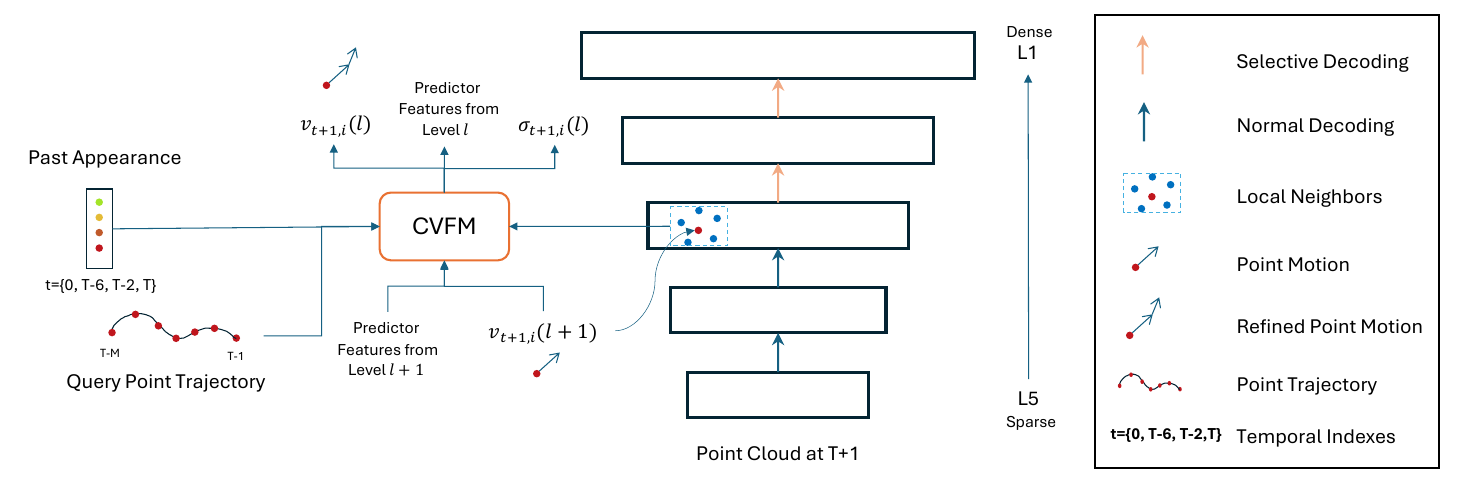}

   \caption{\textbf{Long-term Tracking Framework}. Given a sequence of point clouds as input, 
 we use a U-Net based backbone to extract the point cloud's features hierarchically. For simplicity, only the decoder branch is shown. At each level, we refine the sparse motion from the previous level of the backbone  for the query point  by jointly considering multiple past appearances and the past motion of the query. The motion predicted at level $1$ is used as the final motion of the query point from frame t to t+1.}
 \vskip -0.15in
   \label{fig:overall_framework}
\end{figure*}
We aim to track any point in a video of a dynamic 3D scene. We assume camera poses and depth information have been obtained (e.g. from a SLAM system~\cite{mur2017orb}). With these, the video is converted to a sequence of point clouds $V=\{p_0, \hdots , p_T\}$ where $p_t=\{p_{t,i}\}$, and $p_{t,i} \in R^{3}$ denotes the 3D coordinates of each point $i$ in the point cloud at time step $t$. Let $q_{t,j} \in R^3$ be the $j^{th}$ query point at $t$. For each query point $q_{t,j}$, the model predicts the 3D motion $v_{t+1,j} = q_{t+1,j}-q_{t,j}$ and the occlusion status $\sigma_{t+1,j}\in \{0,1\}$ in the next frame. %The predicted query point's location $q_{t+1,j}$ is then fed into the model, and %
The same process is then repeated autoregressively for long-term point tracking.

The multi-level features for each scene point are obtained using a point-based U-Net backbone \cite{wu2020pointpwc}, comprising an encoder and a decoder. Let $p_t(l)$, $f_t^{p,E}(l)$, and $f_t^{p,D}(l)$ represent the point cloud used at level $l$ and its corresponding encoder/decoder features extracted at level $l$ from our backbone where $l=1$ represents the densest level and $l=L$ the sparsest level. Here, $p_t(l)$ is derived by applying grid-subsampling on $p_t(l-1)$ with $p_t(1) = p_t$. For simplicity, unless explicitly stated, we illustrate the algorithm on a single level and hence refer to the point cloud and their features as $p_t$ and $f_t^{p}$. Unless specified, the features $f_t^p$ represent the decoder features $f_t^{p,D}$. We use $f_{t,i}^p$ to denote the decoder feature at a specific point location $p_{t,i}$.

% \subsection{Background}
% \redtext{Shorten this background section}
% \subsection{Motion Estimation - From 2-Frame to Multi-Frame}
\subsection{Background - PointPWC}
% TODO: SHORTEN THIS SUBSECTION ABOUT POINTPWC
PointPWC~\cite{wu2020pointpwc} is a deep  network that can be trained in either a supervised or unsupervised fashion to predict the scene flow between two point clouds. In PointPWC, a patch-to-patch strategy is used to obtain a robust cost volume and increase the receptive field.
\begin{multline}
     C(p_{t,i}) = \sum_{p_{t,u}\in N_t(p_{t,i})} W_t(p_{t,i}, p_{t,u}) \\
     \sum_{p_{t+1,j}\in N_{t+1}(p_{t,u})} W_{t+1}(p_{t,u}, p_{t+1,j}) cost(p_{t,u}, p_{t+1,j})
     \label{eq:ppwc_cost}
\end{multline}
\begin{equation}
    W_t(p_{t,i}, p_{t,u}) = MLP(p_{t,i}-p_{t,u}) \nonumber
\end{equation}
%     C(p_{t,i}) = \sum_{p_{t,u}\in N_t(p_{t,i})} W_t(p_{t,i}, p_{t,u})\sum_{p_{t+1,j}\in N_{t+1}(p_{t,u})} W_{t+1}(p_{t,u}, p_{t+1,j}) cost(p_{t,i}, p_{t+1,j}).
% \end{equation}
where $N_t(p)$ represents a neighborhood around $p$ at time $t$, and $cost(p_{t,u}, p_{t+1,j})$ refers to the matching cost between two points. The cost is computed via MLP using the concatenation of color features at $p_{t,u}$ and $p_{t+1,j}$ and the relative position between the points as input~\cite{wu2020pointpwc}:
\begin{equation}
    cost(p_{t,u}, p_{t+1,j})=MLP([f^{p}_{t,u}, f^{p}_{t+1,j}, p_{t+1,j}-p_{t,u}]). \nonumber
\end{equation}
The idea of the cost volume is to incorporate matching between points in the neighborhood $N_{t}(p_{t,i})$ and the neighbors of each point in the frame $t+1$, which is similar to a small 2D window that is commonly used to derive 2D cost volumes~\cite{sun2018pwc} -- it  increases the range that points $p_{t,i}$ can match to as well as the robustness of the matching.

Given the cost volume $C(p_{t,i})$ of a point $p_{t,i}$ at level $l$, the PointPWC framework estimates the flow for each point in a coarse to fine manner. Specifically, given the predictor features from level $l+1$, the framework first upsamples them to points on level $l$, then concatenates them with the cost volume $C(p_{t,i})$, as well as the decoder feature $f_{t,i}^p$ at level $l$. These are used together by the flow predictor to generate the predictor features and the residual scene flow at level $l$. The latter is then summed with the upsampled scene flow from level $l+1$ as the predicted flow at level $l$.

\subsection{Cost Volume for Long-Term Point Appearance}
% To facilitate long-term point tracking, we calculate the cost volume using a simpler point-to-patch method instead.

One difference between our long-term point tracking framework and scene flow is the existence of query points that are not necessarily within the given point cloud. Given the hierarchical decoder features of the point cloud, $f^{p}_{t}$, the feature of the query point $q_{t,i}$ can be extracted by applying a PointConv layer as follows:
\begin{equation}
    % f^{q,D}_{t,i}(l) = PC(\phi_{t,i}, \{p_{t,j}(l)\in N_q(\phi_{t,i}(l)), f^{p,D}_{t,j}(l) \})
    f^{q}_{t,i} = MLP\left(\sum_{p_{t,j}\in N(q_{t,i})} W(q_{t,i}, p_{t,j}) f^{p}_{t,j}\right). \label{eq:interpolation}
\end{equation}

Since a single point does not contain enough appearance information, when talking about the appearance of each query point $q_i$, it should be understood as the appearance of the local region containing that point, which could deform from time to time. By jointly considering multiple  appearances,  tracking performance can be improved. Specifically, given a set of appearances of a query point $q_i$ up to the time step $t$, $F^q_{t,i} = \{f^{q}_{t_1,i}, f^{q}_{t_2,i}, \hdots, f^{q}_{t_n,i}\}$, where $t_1,t_2,...,t_n$ represent the frames storing the query's appearances ($t_n \leq t$. Refer to Sec. \ref{sec: long-term tracking module}), we can obtain a set of cost volumes $C^q_{t,i} = \{C_{t_1}(q_{t,i}), \hdots, C_{t_n}(q_{t,i})\}$ as follows:

\begin{equation}
    C_{t_k}(q_{t,i}) =\sum_{p_{t+1,j}\in N(q_{t,i})}W(q_{t,i},p_{t+1,j}) cost_{t_k}(q_{t,i}, p_{t+1,j})
    \label{eq: cost volume}
\end{equation}
\begin{equation*}
    cost_{t_k}(q_{t,i}, p_{t+1,j}) = MLP([f^{q}_{t_k, i}, f^p_{t+1,j}, p_{t+1,j}-q_{t,i}])
\end{equation*}
where we use the feature at timestep $t_k$ to account for the query's multiple appearances.

Note that here we used a simpler cost volume construction without the patch-to-patch formulation as in Eq. (\ref{eq:ppwc_cost}). This is because for long-term tracking, it is difficult to define neighbors at frame $t+1$ from a past frame $t_k$ that could be very far apart temporally. Results in Table \ref{tab:scece_flow_est} show that our cost volume formulation is only slightly less effective than the patch-to-patch formulation. %, it reduces memory consumption significantly as well. %With the larger point clouds we can now fit in memory, our model can still rival other scene flow methods and outperform the original PointPWC in Sec.~\ref{sec: scene-flow exp}.

% where $\phi_{t,i}$ is the current position of the query point $q_i$ at time $t$, $N_q(\phi_{t,i})$ is the neighborhood of the query point $i$.
Another important aspect of the query point features is that based on Eq. (\ref{eq:interpolation}), they are only affected by scene points surrounding the query point at the same level. Therefore, we propose to \textbf{selectively decode} only the points surrounding the query points and prune other points to reduce total computation and memory consumption, especially during training time. We only use this selective decoding strategy for $l \in \{1, 2\}$, which are the two densest levels and thus have the highest decoder memory requirements. By utilizing selective decoding, we increased the number of points per frame that fit into GPU memory from $8,192$ to $60,000$ (with 16 frames used for each mini-batch during training), significantly improving the algorithm's performance, particularly its capability of making predictions with sub-pixel accuracy. For scene flow, this recovered the lost ground we had with the simpler cost volume formulation (Table~\ref{tab:scece_flow_est}), and ablations for long-term point tracking can be found in the supplementary materials.

\subsection{Fusion of Appearance and Motion Cues}
\label{sec: long-term tracking module}
\begin{figure*}
  \centering
   \includegraphics[width=1.\linewidth]{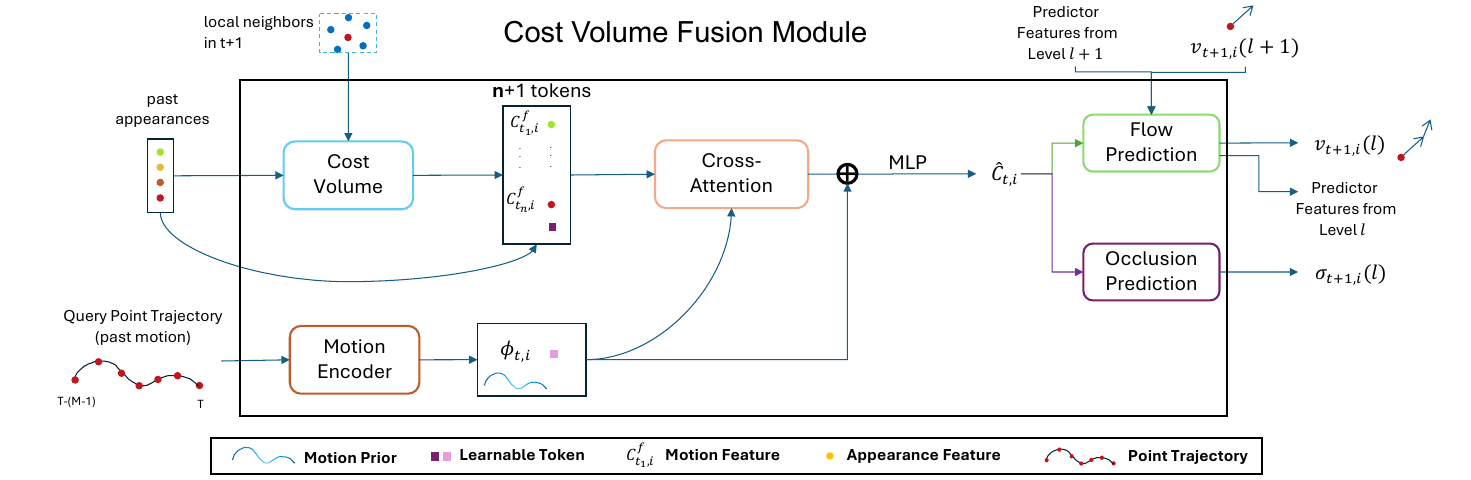}

   \caption{\textbf{Cost Volume Fusion Module}. We propose a novel Cost Volume Fusion Module to predict the query point motion by jointly considering multiple appearances and the past motion trajectory of the query. These appearances are used to compute a set of cost volumes, which are combined with the motion prior via cross-attention in the transformer layer, followed by an MLP. The output features from the MLP are subsequently used to predict the refined motion and the occlusion status of the query point.}
   \label{fig:motion_reasoning}
   \vskip -0.15in
\end{figure*}

A major benefit of long-term point tracking over simply chaining 2-frame scene flows is the capability to incorporate motion cues. Motion cues can help produce motion estimates during occlusion or a blurry frame. However, due to our goal of  non-rigid point tracking, points can move in surprising and different ways that cannot be easily captured by a motion prior. In those cases, a good appearance tracking module should take over and predict more precise locations. 

To properly consider multi-frame appearance and motion jointly, we devise a novel data-driven Cost Volume Fusion module that softly combines motion-based and appearance-matching-based features. The output of this combination is then used to predict the actual motion and occlusion status for each point in the target frame. Below, we detail the specific components of the module.

%\subsubsection{Motion Prior}

% The quality of the matching process heavily depends on the appearance feature of each query point. 
% \subsubsection{Multimodality Multi-Frame Motion Prediction}
\subsubsection{Cost Volume Fusion Module} 
To provide the motion prior for the network, the last $M$ predicted motions of the query point, $v_{(M,t), i}=[v_{t-(M-1),i}, \hdots, v_{t,i}]$, are concatenated and encoded with an MLP 
%including a linear layer % REMOVE AS SUGGESTED
followed by a group normalization layer~\cite{GroupNorm2018} to obtain a motion prior vector $\phi_{t,i}=MLP(v_{(M,t), i})$.
% Here, $\Delta\phi_{t_j,i}=\phi_{t_j+1,i}-\phi_{t_j,i}$, representing the motion of the query point from frame $t_j$ to frame $t_j+1$. 
%$\overrightarrow{a}$
% $\Delta(x,y,z)_t$
Note that, in the beginning of the video, the list of past motions $v_{(M,t), i}$ is initialized with zeros. Additionally, each level $l$ utilizes a separate MLP to encode the motion prior while using the same list of past motions as input.

For each appearance feature of the query extracted at $t_k$, we calculate the corresponding cost volume $C_{t_k,i}$ (a simplified notation for $C_{t_k}(q_{t,i})$) using Eq. (\ref{eq: cost volume}). Each of these cost volumes contains appearance-matching-based motion information. In the experiment, to estimate the motion of a query point from $T$ to $T+1$, we extract the query appearance from the following frames, $\{0, T-6, T-2, T\}$ to reduce the appearance redundancy from consecutive frames.

The cost volume module is shown in Fig.~\ref{fig:motion_reasoning}.
To predict the motion of the $i^{th}$ query in the current frame at the sparse level $l$, our network jointly uses the motion and appearance information extracted from multiple time steps in the past. Specifically, unlike approaches for short-term point correspondence or scene flow, we maintain a list of potential appearance vectors for each query point as mentioned above and calculate its corresponding cost volume with Eq. (\ref{eq: cost volume}).
%For each appearance vector, we calculate the corresponding cost volume feature~\cite{wu2020pointpwc}, $C_{t_j,i}$. Each of these cost volume features contains appearance-matching-based motion information. % MOVE TO THE PREVIOUS PARAGRAPH AS SUGGESTED

Besides, we also consider pure motion-based information from our motion prior $\phi_{t,i}$. This information is especially beneficial when the query point is not visible, as well as other cases where the appearance is ambiguous. %and cannot be matched well. %, resulting in noisy appearance-matching-based motion information. 

% \redtext{rewrite this part}
Each cost volume $C_{t_k,i}$ is concatenated with the corresponding point appearance $f^q_{t_k,i}$ to obtain $C^f_{t_k,i}$. It contains information about the matching between  points from frame $t+1$ and past query point appearances. The motion prior $\phi_{t,i}$ is concatenated with a learnable feature token. We propose to use the motion prior as the query and to cross-attend to all the $\{C^f_{t_k,i}\}_{k=1}^n$. The intuition is to select the points that match some past query point appearance while are also plausible w.r.t. the motion trajectory. Besides, we also use a learnable token $E$ to allow the module to rely on the motion prior when no appearance-matching information is available due to occlusion.
This module can be implemented by stacking multiple transformer decoder~\cite{vaswani2017attention} layers together.
% Finally, a sequence of $M+1$ tokens from above feature vectors is constructed as $T^{C,\phi}_{t,i}=\{C^f_{t_1,i},\hdots, C^f_{t_M,i}, \phi^{MP,f}_{t,i}\}$. 
% A transformer encoder layer followed by an MLP are used to predict the combined point feature and the motion feature as follows:
\begin{align}
    O_{t,i} &= \text{Cross-Attn}(\{C^f_{t_1,i},\hdots, C^f_{t_n,i}, E\}, \phi_{t,i}) \nonumber \\
    \hat{C}_{t,i} &= MLP(O_{t,i}+\phi_{t,i}) \nonumber
\end{align}

Finally, to predict the motion at the current level $l$, $v_{t,i}$, we use the flow predictor head from \cite{wu2020pointpwc} which takes the transformer features $\hat{C}_{t,i}$ and the predictor features from level $l+1$ as input. %The predictor then produces the residual motion and adds it to the predicted motion at level $l+1$ to obtain $v_{t,i}$.

Besides, the transformer features $\hat{C}_{t,i}$ are fed to an MLP to predict the occlusion status at each level $l$. By training this, we encourage the model to store the occlusion information within the cost volumes.
However, during inference, we only use the predicted occlusion at level $1$ as the final occlusion prediction for each query point.

\subsection{Model Training}
% \redtext{Mention the tracking and scene flow losses}
Our model relies on the estimated frame-to-frame motion of each point to construct the long-term point trajectory. Therefore, instead of directly training the model on the long-term tracking data, we split the training process into two stages:
\begin{itemize}
    \item  \textbf{Scene flow pretraining}. The whole model is pre-trained with the scene flow datasets. Each training sample includes two consecutive frames randomly sampled from a video. After the training, the model can achieve competitive performance in the scene flow estimation task. 
    \item \textbf{Long-term tracking}. In this stage, the Cost Volume Fusion Module is added  to handle multiple appearances of the query point and its past trajectory. The network is trained with randomly sampled longer videos.
\end{itemize}
By following this two-stage training pipeline, we can utilize synthetic scene flow datasets to improve the overall tracking performance and stabilize the training process.

We supervise the model by the GT point position and the GT scene flow as follow:
\begin{equation}
    L^{track} = \frac{1}{Tn_q}\sum_{l=1}^L \sum_{t=1}^T \sum_{i=1}^{n_q} \alpha^{l-1}  |q_{t,i}-\hat{q}_{t,i}|_1 \nonumber
\end{equation}
% Besides the query points, we also supervise the motion of all the scene points:
\begin{equation}
    L^{sf} = \frac{1}{T}\sum_{l=1}^L \sum_{t=1}^T \sum_{i=1}^{|p_t|}  \gamma^{l-1} |\Delta p_{t,i}-\Delta\hat{p}_{t,i}|_1 \nonumber
\end{equation}
where $q_{t,i}$ and $\hat{q}_{t,i}$ are the predicted and the ground truth positions, and $\Delta p_{t,i}$ and $\Delta\hat{p}_{t,i}$ are the predicted  and the ground truth flow of the scene point $p_{t,i}$. %\redtext{REMOVE THIS LATER: Do not change $\Delta p_{t,i}$ to $v_{t+1,i}$. }

Motion in the 3D world is usually smooth in terms of both direction and magnitude unless the target is affected by external force from a collision. To encourage such smoothness property, we introduce a smoothness loss that minimizes the difference between the predicted motions in consecutive frames of each query point.
% Let's $m^i_t(l)$ be the predicted motion of the $i^{th}$ query point from frame $t$ to $t+1$ at the $l^{th}$ level. 
The motion smoothness can be defined over all query points as follow:
\begin{equation}
    L^{smooth}=\frac{1}{LTn_q}\sum_{l=0}^L\sum_{i=1}^{n_q}\sum_{t=0}^{T-1}||v_{t,i}-v_{t+1,i}||_1
\end{equation}

We also attempted to use $L^{rigid}$ and $L^{iso}$, the rigidity and isometry losses from ~\cite{luiten2023dynamic}. 
Altogether, we train the model using the weighted sum of the above losses:
\begin{equation}
    L = \lambda_1 \cdot L^{sf} + \lambda_2 \cdot L^{track} + \lambda_3 \cdot L^{smooth} + \lambda_4 \cdot L^{rigid} + \lambda_5 \cdot L^{iso} \nonumber
\end{equation}
We use grid-search to find the optimal values for these hyper parameters. During the scene flow pretraining stage, we only use $L^{sf}$ and $L^{track}$. $\lambda_1$ and $\lambda_2$ are set to 2 and 1 respectively. During the second stage, we set $\lambda_1=2$, $\lambda_2=1$, $\lambda_3=0.3$, and $\lambda_4=\lambda_5=0.2$.

\section{Experiments}
\subsection{Dataset and Training Details}
We use the FlyingThings~\cite{mayer2016large} dataset to pre-train the scene flow model, and then train and test on two separate datasets, TapVid-Kubric~\cite{doersch2022tap} and PointOdyssey~\cite{zheng2023pointodyssey}. \\
\noindent \textbf{TapVid-Kubric} is a synthetic dataset with $9,760$ training videos and 250 testing videos. Each video has 24 frames with resolution $256\times256$. In each validation video, 256 query points are randomly sampled from all the frames. The model is required to track these points in the rest of the video. In the training set, we can generate an arbitrary number of query points for training. Because the dataset is synthetic, we can generate ground truth depth for all points.
%\subsection{Training}

We first build a scene flow task based on the point tracking ground truth on TapVid-Kubric's training split and fine-tune the pretrained model on this task. %fine-tune the pretrained model on the TapVid-Kubric-based \footnote{We build the scene flow dataset based on the point tracking ground truth of TapVid-Kubric's training split.} scene flow datasets. 
Then, with the encoder and decoder backbone frozen to save GPU memory, the full model is fine-tuned on 16-frame videos from the point tracking dataset constructed from TapVid-Kubric~\cite{doersch2022tap}. For data augmentation, we use random horizontal flipping, random scaling the point cloud coordinates, and random temporal flipping. We use a batch size of 16 during the scene flow pre-training stages and a batch size of 8 during the training of the full model.

\noindent \textbf{PointOdyssey} %In addition to TapVid-Kubric, Point Odyssey dataset 
provides much longer synthetic videos (over 1,000 frames and up to 4,000 frames) for training and testing. The dataset includes 131/15/13 videos in the training/validation/testing split.
%For longer videos, we use PointOdyssey dataset. \redtext{Introduce about Point Odyssey. Why is it important}
%\subsection{Testing}
%\redtext{Talk about training on Point Odyssey because it's slightly different.}
Because Point Odyssey does not provide scene flow ground truth, we augment a single frame with random translations and rotations to simulate scene flow data. The model is first fine-tuned on this simulated scene flow data before being trained on the entire training set.
We observe that the model supervised with the simulated scene flow data tends to converge faster in the second phase than the one trained with self-supervised loss.

\noindent \textbf{Backbone} We utilized a U-Net-based PointConvFormer~\cite{wu2023pointconvformer} backbone. This is similar to PCFPWC in Table \ref{tab:scece_flow_est} but with our simplied cost volume (Eq.~(\ref{eq: cost volume})) instead of theirs (Eq.~(\ref{eq:ppwc_cost})).

\noindent \textbf{Inference}. During the inference stage, we have query points that can appear in any place in the video. Hence, we run the model twice (forward and backward) for each video to track the query points in both directions.

\subsection{Metrics}
We extend previous 2D point tracking metrics to 3D:
\begin{itemize}
\vspace{-0.03in}
    \item Occlusion accuracy (OA) is the accuracy of the occlusion prediction for each query point on each frame.
    \vspace{-0.05in}
    \item $\delta^x$ measures the position accuracy of the predicted point on each frame where the point is visible. A predicted point is considered correct if it is within $x$ centimeters (\textit{cm}) from the ground truth position. %To normalize the distance from the prediction to the ground truth, each frame is resized to $256\times256$.
    \vspace{-0.05in}
    \item $\delta^{avg}$ is the average of $\delta^x$ with $x\in[1, 2, 4, 8, 16]$  (\textit{cm}).
\vspace{-0.03in}
\end{itemize}
%, hence we define and use the 3D metrics.
%pecifically, we also use the position accuracy as the main metric with the distance $x\in [1, 2, 4, 8, 16]$ centimeters ($cm$) and 
For PointOdyssey where the videos are  longer, we also adopt the survival rate metric~\cite{luiten2023dynamic} (SR) which is the average number of frames before each tracked point drifts $T$ $cm$ away from the ground truth position, divided by the number of frames in the video, with $T =50$.

We also report results with 2D metrics in order to compare with other 2D methods, but those are secondary results because the primary goal of this paper is 3D tracking.%, we project the predicted 3D points back to the image space.
%Then the model is evaluated using the following metrics:

% \redtext{Back project to 3D. Mention 3d metrics in PIPS+}

% To compare with other 2d baselines, we first run the baselines on RGB images to obtain the predicted 2d coordinates. These 2d coordinates are lifted to 3d using the known camera pose. The depth of each point is obtained by interpolation on the provided depth map.
% \subsection{Scene flow estimation}
% We show in Tab.\ref{tab:scece_flow_est} that our model can achieve competitive performance in scene flow estimation task on the FlyingThing dataset.

\subsection{Scene Flow Pre-Training Results}
\label{sec: scene-flow exp}
In Table~\ref{tab:scece_flow_est}, we show the scene flow pretraining results on the FlyingThings dataset. Our framework outperforms PointPWCNet and other 2-frame baselines. Our scene flow performance was significantly improved when we used an input point cloud size of $60,000$ points over the conventional $8,000$ points used in \cite{wu2020pointpwc} despite our simpler cost volume computation than PointPWCNet and PCF-PWCNet. 
\begin{table}
    \centering
    \begin{tabular}{|c|c|}
        \hline
        \textbf{Methods} & \textbf{EPE3D(m)}$\downarrow$ \\ \hline
        PointPWC~\cite{wu2020pointpwc} & 0.0588\\ \hline
        PCFPWC~\cite{wu2023pointconvformer} & \bluetext{\textbf{0.0416}}\\ \hline
        PV-RAFT~\cite{wei2021pv} & 0.0461\\ \hline
        FLOT~\cite{puy2020flot} & 0.0520\\ \hline
        HCRF-Flow~\cite{li2021hcrf} &0.0488 \\ \hline
        HPLFlowNet~\cite{gu2019hplflownet} & 0.0804\\ \hline
        FlowNet3D~\cite{liu2019flownet3d} & 0.1136 \\ \hline
        \textbf{Ours - 8k points} & 0.0509\\ \hline
        \textbf{Ours - 60k points}  & \redtext{\textbf{0.0399}}\\  \hline
    \end{tabular}
    \vskip -0.05in
    \caption{Scene flow estimation on the FlyingThing dataset}
    \label{tab:scece_flow_est}
 \vskip -0.2in    
\end{table}

\begin{figure*}[htb]
% \vskip -0.2in
  \centering
   \includegraphics[width=.8\linewidth]{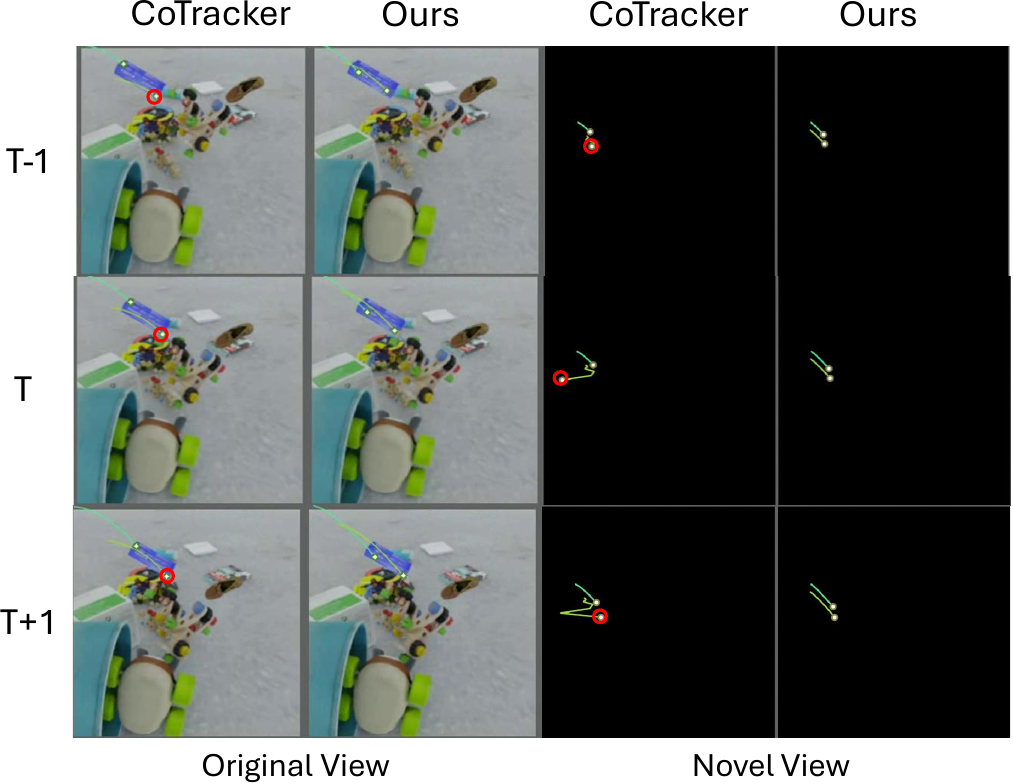}
% \vskip -0.05in
   \caption{\textbf{Qualitative Results}. We reproject the results of CoTracker into 3D and back-project that into a different view point. One can see that because of small errors in 2D leading the CoTracker result on the red circled point off the blue object at time $T$, it incurs significant 3D error which can be seen as a sudden jump in the trajectory if rendered from a novel viewpoint. (Best viewed in color)}
   \label{fig:qualitative-long-trajs}
   \vskip -0.05in
\end{figure*}

\subsection{ 3D \& 2D Evaluations}
To the best of our knowledge, no prior deep learning-based work tackled the problem of generalizable 3D point tracking. Hence, we mainly compare with 2D baselines and simple scene flow chaining. Since the dataset is synthetic, we can lift 2D tracking results to 3D using the ground truth camera pose and depth map.
%We lift the results of Tapir~\cite{doersch2023tapir} and CoTracker~\cite{karaev2023cotracker} to 3d using the provided camera pose and the depth map.
Due to the subpixel-level predictions from the 2D point trackers, the depth of each point is obtained by interpolation on the provided depth map. For the occluded points, the depth is linearly interpolated using the depth of that point before and after the occlusion. Results with alternative nearest neighbor interpolation are similar and are shown in the supplementary material.

\begin{table}[b]
% \vskip -0.1in
\centering
\scalebox{0.9}{
\begin{tabular}{|c|c|c|c|}
\hline
\
\textbf{}                 & \textbf{OA} &   \textbf{3D - $\delta^{avg}$} & \textbf{3D -} $\delta^{avg}_{occluded}$ \\ \hline
% PIPs                       &  &  & & \\ \hline
TAPIR                      & \redtext{\textbf{96.5}} &  46.9 (\textbf{-26.2}) & 3.8 \\ \hline
CoTracker                  & 92.5  &  \bluetext{\textbf{57.8}} (\textbf{-15.3}) &  \bluetext{\textbf{8.7}}\\ \hline
Scene Flow Chaining & - & 45.6 & 5.6 \\ \hline
\textbf{Ours}                       & \bluetext{\textbf{93.4}}  &  \redtext{\textbf{73.8}} & \redtext{\textbf{44.4}}\\ \hline
\end{tabular}}
\vskip -0.05in
\caption{Results on 3D Kubric. Best results are shown in red and second best in blue. (Best viewed with color)}
\label{tab:3d_results}
\vskip -0.15in
\end{table}

\begin{table}[b]
\begin{tabular}{|c|c|c|c|c|c|c|}
\hline
       & \multicolumn{2}{c|}{\textbf{128 Frames}}             & \multicolumn{2}{c|}{\textbf{512 Frames}}             & \multicolumn{2}{c|}{\textbf{Full Seq}}            \\ \hline
       & SR   & $\delta^{avg}$ & SR   & $\delta^{avg}$ & SR   & $\delta^{avg}$ \\ \hline
PIPS++~\cite{zheng2023pointodyssey} & 64.4 & 35.0         & 39.7 & 26.4         & 16.0 & 14.0         \\ \hline
% CoTracker & 68.7 & 36.1 & 44.3 & 25.2  & 17.9 & 10.8\\ \hline
\textbf{Ours}   & \redtext{\textbf{91.0}} & \redtext{\textbf{65.1}}         & \redtext{\textbf{82.6}} & \redtext{\textbf{52.0}}         & \redtext{\textbf{68.5}} & \redtext{\textbf{35.0}}         \\ \hline
\end{tabular}
\vskip -0.05in
\caption{Results on 3D Point Odyssey.}%*SR: survival rate with the threshold at 50 cm. }
%\redtext{COTRACKER IS TRAINED ON KUBRIC. NO POINTODYSSEY CKPT RELEASED.}}
\label{tab:3d_pod}
\end{table}

\begin{table}[b]
% \vskip -0.1in
\centering
\scalebox{0.8}{
\begin{tabular}{|c|c|c|c|c|}
\hline
\
\textbf{}                 & \textbf{OA} & \textbf{2D - $\delta^{avg}$} & \textbf{2D - AJ} & \textbf{2D -} $\delta^{avg}_{occluded}$ \\ \hline
COTR~\cite{jiang2021cotr}                       & 78.55 & 60.7 & 40.1 &  -\\ \hline
RAFT~\cite{teed2020raft}                       & 86.4 & 58.2 & 41.2 & -\\ \hline
PIPs~\cite{harley2022particle}                       & 88.6 & 74.8 & 59.1 & -\\ \hline
Tap-Net~\cite{doersch2022tap}                     & 93.0 & 77.7 & 65.4 & -\\ \hline
TAPIR* (1 iter)~\cite{doersch2023tapir}                      & 94.6 & 88.8  & 81.0 & 27.5 \\ \hline
CoTracker* (1 iter)~\cite{karaev2023cotracker}                  & - & 90.0  & - & 56.9 \\ \hline
Scene Flow Chaining &- & 77.0  & - & 25.8 \\ \hline
% \textbf{Ours }                      & \bluetext{\textbf{93.4}} & 87.0 & 74.4 & \bluetext{\textbf{60.8}}\\ 
\textbf{Ours }                      & \bluetext{\textbf{93.4}} & 87.8 & 75.4 & \bluetext{\textbf{61.2}}\\ 
\hline
\midrule
TAPIR*~\cite{doersch2023tapir}                      & \redtext{\textbf{96.5}} & \bluetext{\textbf{92.9} }  & \redtext{\textbf{86.1}} & 36.8 \\ \hline
CoTracker*~\cite{karaev2023cotracker}                  & 92.5 & \redtext{\textbf{93.9}}  & \bluetext{\textbf{84.5}} & \redtext{\textbf{66.0}} \\ \hline

\end{tabular}}
\vskip -0.05in
\caption{Results on 2D Kubric. *: reproduced results. Best results are shown in red and second best in blue. TAPIR and CoTracker used an hourglass network that re-runs the encoder and decoder several times, whereas our method does not use hourglass and is comparable with their results with a single iteration}
\label{tab: 2d-results}
\end{table}

\begin{table*}[ht]
% \vskip -0.1in
\centering
\begin{tabular}{|c|c|c|c|c|c|}
\hline
\textbf{}                 & \textbf{OA} & \textbf{2D - $\delta^{avg}$} & \textbf{2D - $\delta^{avg}_{occluded}$} & \textbf{3D - $\delta^{avg}$} & \textbf{3D - $\delta^{avg}_{occluded}$} \\ \hline
Multi-App + Motion Prior  & \redtext{\textbf{93.4}} & \redtext{\textbf{87.0}}  & \redtext{\textbf{60.8}}  & \redtext{\textbf{73.1}}  & \redtext{\textbf{44.0}}  \\ \hline
Multi-Appearance                 & 92.9 & 86.0 & 57.9  & 71.6  & 40.5  \\ \hline
% Single-App + Motion Prior & 93.7 & 87.0 & 61.4 & 73.3 & 44.9 \\ \hline
Single-Appearance               & 91.3 & 83.5 & 52.7 & 61.6 & 29.8 \\ \hline
\end{tabular}
\vskip -0.05in
\caption{Ablation on Appearance and Motion Priors}
\label{tab: ablation-motion-app}
\vskip -0.05in
\end{table*}
\begin{table*}[ht]
\centering
\begin{tabular}{|c|c|c|c|c|c|}
\hline
\
\textbf{}                 & \textbf{OA} & \textbf{2d} - $\delta^{avg}$ & \textbf{2d} - $\delta^{avg}_{occluded}$ & \textbf{3d - $\delta^{avg}$} & \textbf{3d} - $\delta^{avg}_{occluded}$ \\ \hline
Ours & 93.4  & 87.0 & 60.8 & 73.1 & 44.0\\ \hline
Rigid \& Isometry & 93.4 & 86.8 & 60.9 & 73.1 & 44.4 \\ \hline
Smoothness & 92.4 & 83.0 & 55.3 & 55.3 & 30.0 \\ \hline
                            
\end{tabular}
\vskip -0.05in
\caption{Ablation on Regularizations}
\label{tab:ablation-regularization}
\vskip -0.2in
% \vskip -0.3in
\end{table*}
%\redtext{What more?}
%While our method outperforms CoTracker~\cite{karaev2023cotracker} and Tapir~\cite{doersch2023tapir} in 2D space by 5.9\% and 6.9\% respectively under $\delta^{avg}$ (Table~\ref{tab: 2d-results}), they lag behind by more than 15\% and 26\% respectively in 3D. 

Table \ref{tab:3d_results} shows 3D point tracking results comparing our proposed approach with state-of-the-art 2D approaches TAPIR~\cite{doersch2023tapir} and CoTracker~\cite{karaev2023cotracker}, as well as simply chaining the pretrained scene flow. We significantly outperform both approaches by $15.3\%$ and $26.2\%$ points on the $3D- \delta^{avg}$ metric. The performance difference is more significant in the occluded areas, where we record a $44.0\%$ accuracy whereas TAPIR and CoTracker obtain accuracies lower than $10\%$ due to not having a good 3D motion prior to maintain a good track during occlusion. Note that baseline results were already generated by interpolating tracks using the \textbf{ground truth} depth. Better nonlinear interpolation may improve their performance by a bit, but it is unlikely that their performance would catch up to our approach. 

Such significant performance differences support our arguments that even accurate 2D tracking can have significant issues locating accurate tracks in 3D, even with fully known camera pose and ground truth depth. A qualitative illustration in Fig.~\ref{fig:qualitative-long-trajs} indicates the issue with 2D trackers. At frame $T$, although the original view does not indicate a significant error on the red circled point that is being tracked, the tracking actually drifted slightly off the blue object. In consequence, if we render it from a novel viewpoint outside of the original 2D image plane, we can see a significant jump in the trajectory. At time $T+1$, the tracked point went back to the blue object and from the novel view we again see a significant jump in the trajectory. This indicates significant errors in 3D tracking despite low 2D tracking errors. Our approach, on the other hand, works naturally in 3D. Hence it does not suffer from such drifting issues and produces much more consistent 3D tracking results.

Results on the test split of PointOdyssey in Table~\ref{tab:3d_pod} show a similar trend. We measure the Survival Rate and $\delta^{avg}$ on the first 128 frames and 512 frames of each video or the full sequences. Our framework also outperforms the baselines by a large margin across all the metrics. 
%Note that the occlusion accuracy of CoTracker (1 iter) drops significantly because the authors only supervise the prediction in the last step of the refinement process.

In Table \ref{tab: 2d-results}, we show results on the 2D evaluation on Kubric. We projected our 3D results to 2D using the known camera parameters for our approach. Our approach outperformed many baselines and are generally comparable or slightly worse than TAPIR and CoTracker in 2D when only a single iteration is used for them. Note that for our 3D tracking results in Table \ref{tab:3d_results}, we only compared against the full version of TAPIR and CoTracker (i.e., with multiple iterations) and still outperformed them. We did not include the OA and 2D-AJ (Average Jaccard) numbers for CoTracker (1 iter) because CoTracker produces the occlusion status only at the last iteration. TAPIR and CoTracker additionally utilize an hourglass network that decodes, re-encodes and decodes several times to obtain more precise predictions, but that is against the spirit of online algorithms and we did not pursue that path. We did outperform TAPIR significantly in the prediction of occluded points, showing the benefits of interpolation of the motion from 3D.

\subsection{Ablation Experiments}

%\subsubsection{Ablation on Using Multiple Appearances \& Motion Priors}
We analyze the contribution of using multiple appearance features and the motion prior. Results are  presented in Table~\ref{tab: ablation-motion-app}. We demonstrate that by employing multiple appearances and the motion prior, the performance across all 2D and 3D metrics consistently improves over using a single appearance.
% Besides, the improvement under occlusion is also better than that under normal condition. Using multiple appearances of the query point also help the model to withstand the occlusion better. 
% We hypothesize that by incorporating multiple appearances of the query points, the model can better detect the neighboring area of the query point when the object containing it is partially occluded. For that reason, the model can approximate the location of the query point indirectly through the neighboring points.
Additionally, the performance improvement under occlusion surpasses that in normal conditions, showing that utilizing multiple appearances of the query point makes the model more resilient to occlusion. By integrating  appearances from the past, the model can recover from a few bad appearances during or right before/after occlusion and achieve better results. %detect the query point's surrounding area, aiding in locating it indirectly through neighboring points.
% In Table.~\ref{tab: ablation-motion-app}, we a

%Additional ablations, such as those on regularization terms, are shown in the supplemental.
%\subsubsection{Ablation on Regularization Terms}
We also conduct ablation on the regularization terms. Results in Table~\ref{tab:ablation-regularization} show that our smoothness term is very useful in terms of improving the 3D point tracking results. The rigidity and isometry loss from ~\cite{luiten2023dynamic} provide marginal improvements in the 2D results.

% \FloatBarrier
\section{Conclusion}
In this paper, we proposed a 3D long-term point tracking approach based on fusing multiple cost volumes and motion information with a transformer model, which, to the best of our knowledge, is the first generalizable online long-term 3D point tracking approach using deep learning. By selective decoding, we significantly increased the size of the input point cloud that fits into GPU memory, which improves the performance of scene flow and 3D long-term point tracking. In terms of 3D point tracking performance, our approach significantly outperforms scene flow chaining and 2D long-term point tracking approaches even if they are backprojected to 3D with ground truth depths and camera poses, showing the benefits of tracking in 3D. We hope this paper could increase the interest of the community in 3D long-term point tracking. In future work, we plan to utilize our 3D point tracking framework in downstream tasks.
%------------------------------------------------------------------------
% \FloatBarrier

%%%%%%%%% REFERENCES
{\small
\bibliographystyle{ieee_fullname}
\bibliography{egbib}

\begin{thebibliography}{10}\itemsep=-1pt

\bibitem{bozic2020neural}
Aljaz Bozic, Pablo Palafox, Michael Zollh{\"o}fer, Angela Dai, Justus Thies, and Matthias Nie{\ss}ner.
\newblock Neural non-rigid tracking.
\newblock {\em Advances in Neural Information Processing Systems}, 33:18727--18737, 2020.

\bibitem{breu1995linear}
Heinz Breu, Joseph Gil, David Kirkpatrick, and Michael Werman.
\newblock Linear time euclidean distance transform algorithms.
\newblock {\em IEEE Transactions on Pattern Analysis and Machine Intelligence}, 17(5):529--533, 1995.

\bibitem{bupe2024wrong}
Chomba Bupe.
\newblock \url{https://twitter.com/ChombaBupe/status/1764423313164472706}, 2024.

\bibitem{danielsson1980euclidean}
Per-Erik Danielsson.
\newblock Euclidean distance mapping.
\newblock {\em Computer Graphics and image processing}, 14(3):227--248, 1980.

\bibitem{doersch2022tap}
Carl Doersch, Ankush Gupta, Larisa Markeeva, Adri{\`a} Recasens, Lucas Smaira, Yusuf Aytar, Jo{\~a}o Carreira, Andrew Zisserman, and Yi Yang.
\newblock Tap-vid: A benchmark for tracking any point in a video.
\newblock {\em Advances in Neural Information Processing Systems}, 35:13610--13626, 2022.

\bibitem{doersch2023tapir}
Carl Doersch, Yi Yang, Mel Vecerik, Dilara Gokay, Ankush Gupta, Yusuf Aytar, Joao Carreira, and Andrew Zisserman.
\newblock Tapir: Tracking any point with per-frame initialization and temporal refinement.
\newblock {\em arXiv preprint arXiv:2306.08637}, 2023.

\bibitem{fang2024self}
Shaoheng Fang, Zuhong Liu, Mingyu Wang, Chenxin Xu, Yiqi Zhong, and Siheng Chen.
\newblock Self-supervised bird’s eye view motion prediction with cross-modality signals.
\newblock In {\em Proceedings of the AAAI Conference on Artificial Intelligence}, volume~38, pages 1726--1734, 2024.

\bibitem{gu2019hplflownet}
Xiuye Gu, Yijie Wang, Chongruo Wu, Yong~Jae Lee, and Panqu Wang.
\newblock Hplflownet: Hierarchical permutohedral lattice flownet for scene flow estimation on large-scale point clouds.
\newblock In {\em Proceedings of the IEEE/CVF conference on computer vision and pattern recognition}, pages 3254--3263, 2019.

\bibitem{harley2022particle}
Adam~W Harley, Zhaoyuan Fang, and Katerina Fragkiadaki.
\newblock Particle video revisited: Tracking through occlusions using point trajectories.
\newblock In {\em European Conference on Computer Vision}, pages 59--75. Springer, 2022.

\bibitem{jiang2021cotr}
Wei Jiang, Eduard Trulls, Jan Hosang, Andrea Tagliasacchi, and Kwang~Moo Yi.
\newblock Cotr: Correspondence transformer for matching across images.
\newblock In {\em Proceedings of the IEEE/CVF International Conference on Computer Vision}, pages 6207--6217, 2021.

\bibitem{karaev2023cotracker}
Nikita Karaev, Ignacio Rocco, Benjamin Graham, Natalia Neverova, Andrea Vedaldi, and Christian Rupprecht.
\newblock {CoTracker}: It is better to track together.
\newblock 2023.

\bibitem{laptev2005space}
Ivan Laptev.
\newblock On space-time interest points.
\newblock {\em International journal of computer vision}, 64:107--123, 2005.

\bibitem{li2021hcrf}
Ruibo Li, Guosheng Lin, Tong He, Fayao Liu, and Chunhua Shen.
\newblock Hcrf-flow: Scene flow from point clouds with continuous high-order crfs and position-aware flow embedding.
\newblock In {\em Proceedings of the IEEE/CVF Conference on Computer Vision and Pattern Recognition}, pages 364--373, 2021.

\bibitem{li2021neural}
Xueqian Li, Jhony Kaesemodel~Pontes, and Simon Lucey.
\newblock Neural scene flow prior.
\newblock {\em Advances in Neural Information Processing Systems}, 34:7838--7851, 2021.

\bibitem{Li_2023_ICCV}
Xueqian Li, Jianqiao Zheng, Francesco Ferroni, Jhony~Kaesemodel Pontes, and Simon Lucey.
\newblock Fast neural scene flow.
\newblock In {\em Proceedings of the IEEE/CVF International Conference on Computer Vision (ICCV)}, pages 9878--9890, October 2023.

\bibitem{liu2024difflow3d}
Jiuming Liu, Guangming Wang, Weicai Ye, Chaokang Jiang, Jinru Han, Zhe Liu, Guofeng Zhang, Dalong Du, and Hesheng Wang.
\newblock Difflow3d: Toward robust uncertainty-aware scene flow estimation with iterative diffusion-based refinement.
\newblock In {\em Proceedings of the IEEE/CVF Conference on Computer Vision and Pattern Recognition}, pages 15109--15119, 2024.

\bibitem{liu2019flownet3d}
Xingyu Liu, Charles~R Qi, and Leonidas~J Guibas.
\newblock Flownet3d: Learning scene flow in 3d point clouds.
\newblock In {\em Proceedings of the IEEE/CVF conference on computer vision and pattern recognition}, pages 529--537, 2019.

\bibitem{luiten2023dynamic}
Jonathon Luiten, Georgios Kopanas, Bastian Leibe, and Deva Ramanan.
\newblock Dynamic 3d gaussians: Tracking by persistent dynamic view synthesis.
\newblock {\em arXiv preprint arXiv:2308.09713}, 2023.

\bibitem{mayer2016large}
Nikolaus Mayer, Eddy Ilg, Philip Hausser, Philipp Fischer, Daniel Cremers, Alexey Dosovitskiy, and Thomas Brox.
\newblock A large dataset to train convolutional networks for disparity, optical flow, and scene flow estimation.
\newblock In {\em Proceedings of the IEEE conference on computer vision and pattern recognition}, pages 4040--4048, 2016.

\bibitem{mur2017orb}
Raul Mur-Artal and Juan~D Tard{\'o}s.
\newblock Orb-slam2: An open-source slam system for monocular, stereo, and rgb-d cameras.
\newblock {\em IEEE transactions on robotics}, 33(5):1255--1262, 2017.

\bibitem{puy2020flot}
Gilles Puy, Alexandre Boulch, and Renaud Marlet.
\newblock Flot: Scene flow on point clouds guided by optimal transport.
\newblock In {\em European conference on computer vision}, pages 527--544. Springer, 2020.

\bibitem{qi2017pointnet++}
Charles~Ruizhongtai Qi, Li Yi, Hao Su, and Leonidas~J Guibas.
\newblock Pointnet++: Deep hierarchical feature learning on point sets in a metric space.
\newblock {\em Advances in neural information processing systems}, 30, 2017.

\bibitem{schmidt2023tracking}
Adam Schmidt, Omid Mohareri, Simon DiMaio, Michael Yip, and Septimiu~E Salcudean.
\newblock Tracking and mapping in medical computer vision: A review.
\newblock {\em arXiv preprint arXiv:2310.11475}, 2023.

\bibitem{shen2023self}
Yaqi Shen, Le Hui, Jin Xie, and Jian Yang.
\newblock Self-supervised 3d scene flow estimation guided by superpoints.
\newblock In {\em Proceedings of the IEEE/CVF Conference on Computer Vision and Pattern Recognition}, pages 5271--5280, 2023.

\bibitem{sun2018pwc}
Deqing Sun, Xiaodong Yang, Ming-Yu Liu, and Jan Kautz.
\newblock Pwc-net: Cnns for optical flow using pyramid, warping, and cost volume.
\newblock In {\em Proceedings of the IEEE conference on computer vision and pattern recognition}, pages 8934--8943, 2018.

\bibitem{teed2020raft}
Zachary Teed and Jia Deng.
\newblock Raft: Recurrent all-pairs field transforms for optical flow.
\newblock In {\em Computer Vision--ECCV 2020: 16th European Conference, Glasgow, UK, August 23--28, 2020, Proceedings, Part II 16}, pages 402--419. Springer, 2020.

\bibitem{vaswani2017attention}
Ashish Vaswani, Noam Shazeer, Niki Parmar, Jakob Uszkoreit, Llion Jones, Aidan~N Gomez, {\L}ukasz Kaiser, and Illia Polosukhin.
\newblock Attention is all you need.
\newblock {\em Advances in neural information processing systems}, 30, 2017.

\bibitem{wang2021unsupervised}
Guangming Wang, Xiaoyu Tian, Ruiqi Ding, and Hesheng Wang.
\newblock Unsupervised learning of 3d scene flow from monocular camera.
\newblock In {\em 2021 IEEE International Conference on Robotics and Automation (ICRA)}, pages 4325--4331. IEEE, 2021.

\bibitem{wang2023omnimotion}
Qianqian Wang, Yen-Yu Chang, Ruojin Cai, Zhengqi Li, Bharath Hariharan, Aleksander Holynski, and Noah Snavely.
\newblock Tracking everything everywhere all at once.
\newblock In {\em International Conference on Computer Vision}, 2023.

\bibitem{wei2021pv}
Yi Wei, Ziyi Wang, Yongming Rao, Jiwen Lu, and Jie Zhou.
\newblock Pv-raft: Point-voxel correlation fields for scene flow estimation of point clouds.
\newblock In {\em Proceedings of the IEEE/CVF conference on computer vision and pattern recognition}, pages 6954--6963, 2021.

\bibitem{wu2023pointconvformer}
Wenxuan Wu, Li Fuxin, and Qi Shan.
\newblock Pointconvformer: Revenge of the point-based convolution.
\newblock In {\em Proceedings of the IEEE/CVF Conference on Computer Vision and Pattern Recognition}, pages 21802--21813, 2023.

\bibitem{wu2019pointconv}
Wenxuan Wu, Zhongang Qi, and Li Fuxin.
\newblock Pointconv: Deep convolutional networks on 3d point clouds.
\newblock In {\em Proceedings of the IEEE/CVF Conference on computer vision and pattern recognition}, pages 9621--9630, 2019.

\bibitem{wu2020pointpwc}
Wenxuan Wu, Zhi~Yuan Wang, Zhuwen Li, Wei Liu, and Li Fuxin.
\newblock Pointpwc-net: Cost volume on point clouds for (self-) supervised scene flow estimation.
\newblock In {\em Computer Vision--ECCV 2020: 16th European Conference, Glasgow, UK, August 23--28, 2020, Proceedings, Part V 16}, pages 88--107. Springer, 2020.

\bibitem{GroupNorm2018}
Yuxin Wu and Kaiming He.
\newblock Group normalization.
\newblock {\em arXiv:1803.08494}, 2018.

\bibitem{zhang2024seflow}
Qingwen Zhang, Yi Yang, Peizheng Li, Olov Andersson, and Patric Jensfelt.
\newblock Seflow: A self-supervised scene flow method in autonomous driving.
\newblock {\em arXiv preprint arXiv:2407.01702}, 2024.

\bibitem{zhang2024diffsf}
Yushan Zhang, Bastian Wandt, Maria Magnusson, and Michael Felsberg.
\newblock Diffsf: Diffusion models for scene flow estimation.
\newblock {\em arXiv preprint arXiv:2403.05327}, 2024.

\bibitem{zheng2023pointodyssey}
Yang Zheng, Adam~W Harley, Bokui Shen, Gordon Wetzstein, and Leonidas~J Guibas.
\newblock Pointodyssey: A large-scale synthetic dataset for long-term point tracking.
\newblock In {\em Proceedings of the IEEE/CVF International Conference on Computer Vision}, pages 19855--19865, 2023.

\end{thebibliography}
}
\clearpage
\appendix

The supplementary material consists of this document and a video demo. In the document, we provide additional ablation results and result analyses. In the video demo, we show comparisons between our approach and CoTracker, the strongest 2D point tracking baseline used in our experiments. 
%We provide more qualitative results in our videos in the original and the new view (i.e. 45\degree) in the videos.
% The layout of the videos is described in Fig.~\ref{fig: trajectory},

\section{Running Time Breakdown}
In this section, we show the running time breakdown for our model. 
We benchmark our model on the test split of the PointOdyssey dataset. The resolution of each frame in the dataset is $540\times960$. Therefore, the maximum points on each frame is up to  518400. Our average FPS on this dataset is around 2.8.

\begin{figure}[h]
    \centering
    \includegraphics[width=0.7\linewidth]{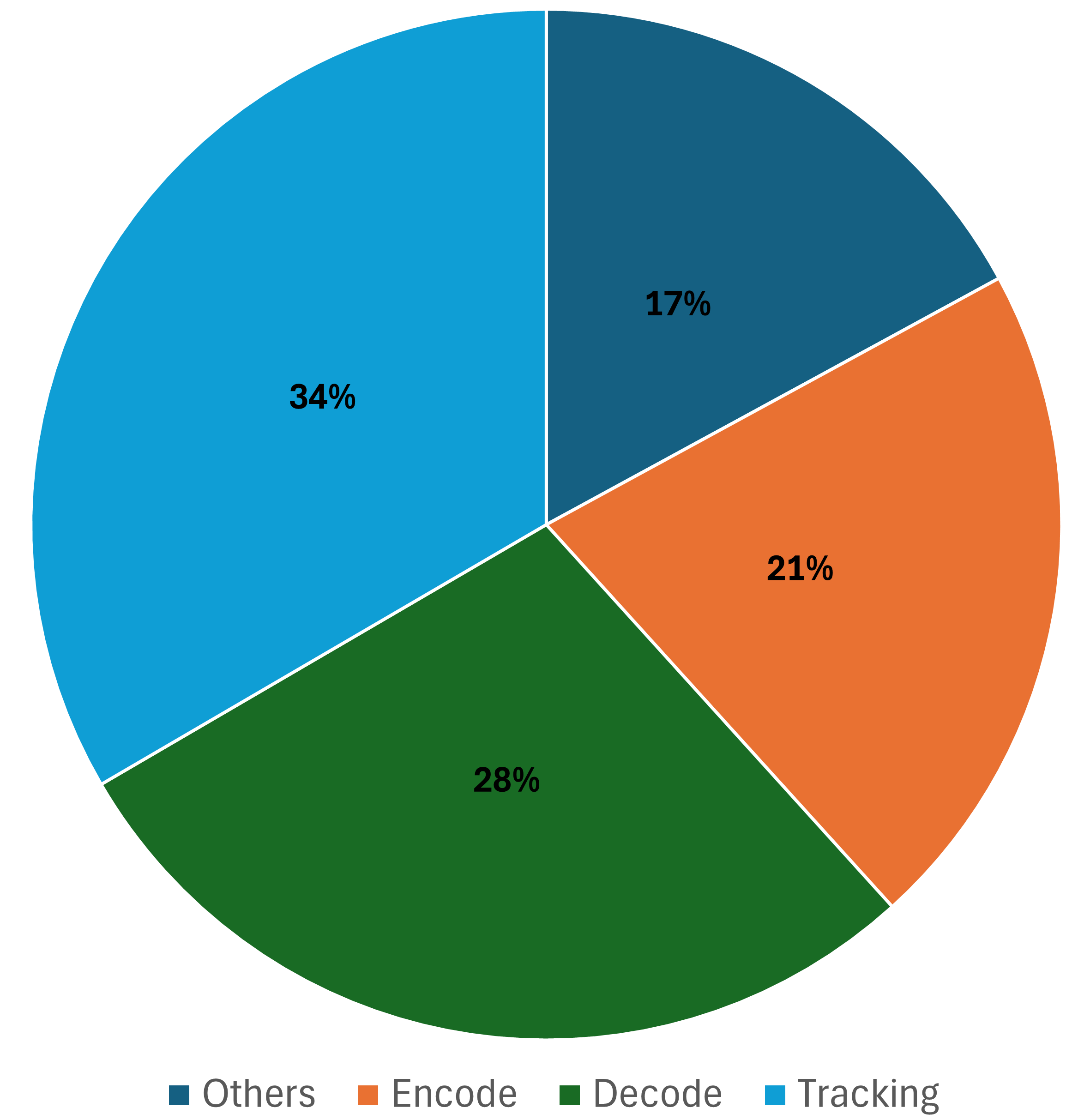}
    \caption{Running Time Percentage by Components}
    \label{fig:time_breakdown}
\end{figure}

As shown in Fig.~\ref{fig:time_breakdown}, our model takes roughly half of the total running to extract the point cloud and query's features (i.e., \textbf{encode} \& \textbf{decode}). The Cost Volume Fusion module, which predicts the query motion, takes roughly one third of the total running time. To further cut down the running time, we can replace our backbone network with a faster network. 
\FloatBarrier
\section{Ablation - Effect of Dense Point Cloud on Long Term Tracking}
As mentioned in the main paper, the selective decoding module enables the use of a dense point cloud by reducing memory consumption significantly. In this ablation, we show the importance of using dense point clouds. As shown in Table.~\ref{tab: dense-sparse}, the overall performance on 
%both short-term (scene flow) and 
long-term tracking is  improved by switching from sparse point cloud input into dense point cloud input. 
%Note that the model we use to evaluate the effect of using a denser point cloud in this ablation does not have the motion prior and is not trained with the rigid and smoothness losses. Consequently, the overall performance is worse than the one reported in the main paper.

\begin{table}[]
\centering
\caption{Ablation on the Effect of Dense Point Cloud in Long-term Tracking}
\label{tab: dense-sparse}
\begin{tabular}{|c|c|c|}
\hline
\textbf{Method}    &  Sparse PC  & Dense PC  \\ \hline
% \textbf{2D} - $\delta^{avg}$  $\uparrow$ & 80.41 & \textbf{82.56}\\ \hline
\textbf{2D} - $\delta^{avg}$  $\uparrow$ & 80.5 & \textbf{87.8} \\ \hline

% Sparse PC & 80.5 & 0.05\\ \hline
% \textbf{Dense PC}  & \textbf{87.8} & \textbf{0.04}\\ \hline

\end{tabular}

\end{table}
% \FloatBarrier
\section{Ablation - Effect of Selective Decoding}
Table~\ref{tab:compression_rate} shows the compression rate we achieve by reducing the number of points to be decoded in the 2 most densest levels. Here, we define the compression rate to be the ratio between the number of points before and after the pruning process of the selective decoding module.

Besides, we can significantly reduce the total memory consumption when training the model - \textbf{Table~\ref{tab: selective-decoding}}. Here, we show memory consumption during the pre-training process on the scene flow dataset with batch size 8. Besides, it also helps to speed up the training process by 1.7 times.
%reduce the overall training time by roughly 1.7 times.
\begin{table}
    \centering
    \caption{Compression Rates (\textbf{CR}) on 2 densest point cloud levels}
    \label{tab:compression_rate}
    \begin{tabular}{|c|c|c|}
        \hline
         & \textbf{Level 1}  & \textbf{Level 2} \\ \hline
        \textbf{CR} & 13.99  & 9.29\\ \hline
    \end{tabular}
    
\end{table}
\begin{table}[]
\centering
\caption{Ablation on the Effect of Using Selective Decoding }
\label{tab: selective-decoding}
\begin{tabular}{|c|c|}
\hline
\textbf{Training Memory}           & \textbf{GBs} \\ \hline
Decode all points                  & 37             \\ \hline
\textbf{Selectively decode points} & \textbf{17}    \\ \hline
\end{tabular}

\end{table}
% \FloatBarrier
\section{In-depth 3D performance comparison to CoTracker}
%We show an example of the tracking in the new viewing direction in Fig.~\ref{fig: trajectory}. More examples can be seen in the video.

% 
% To show the advantage of working directly in 3D, we
While working well in 2D space, CoTracker~\cite{karaev2023cotracker}'s performance significantly drops in 3D. We hypothesize that the main reason is due to noisy depth values used to convert 2D points into 3D. To validate our hypothesis, we visualize 3D tracks using new camera views by rotating the camera around the z-axis as follows. First, we back-project the predicted 2D point tracks of CoTracker into the 3D scene using the original camera pose and the provided depth maps. 
The depth of each predicted query point is obtained by first selecting the four nearest neighbors in the depth map (i.e., 4 depth pixels) and calculating the depth through bilinear interpolation. We also show the results for nearest interpolation in Sec.~\ref{sec: nearest_interp}.
Then, these points are projected to 2D again using new viewing angle.
%viewing directions. 
In contrast to CoTracker, we can directly project our predicted 3D points into the new viewing angles.
%viewing directions. 
The 2D results evaluated in the new camera views are shown in Table.~\ref{tab: camera-rotating}. Although CoTracker performs well in the original view, its performance rapidly declines when these points are projected into new camera views. In contrast, our model exhibits more robustness to changes in viewpoint. %, with only a slight decrease in performance.
% Note that the results of our model are also affected in a smaller degree
% because we do not have the completed 3D point clouds. The \textbf{partial} 3D point clouds that we use are constructed from the depth maps and images from the same viewing angle that CoTracker and Tapir are trained on. 
% In Table.~\ref{tab: camera-rotating}
\begin{table}[]
\centering
\caption{Rotate the camera around the z-axis. The camera always looks at $(0,0,0)$. }
\label{tab: camera-rotating}
\scalebox{0.85}{
\begin{tabular}{|l|c|c|c|c|c|c|c|}
\hline
 & $\delta^{avg}$ & $\delta^{avg}_{15\degree}$ & $\delta^{avg}_{30\degree}$ & $\delta^{avg}_{45\degree}$ & $\delta^{avg}_{60\degree}$ &$\delta^{avg}_{75\degree}$ & $\delta^{avg}_{90\degree}$ \\ \hline
CoTracker      & 93.8 & 87.1        & 81.2        & 74.4        & 65.4        & 52.4        & 43.8        \\ \hline
Ours           & 87.8 & 87.5        & 86.3        & 84.0        & 79.2        & 68.6        & 62.5        \\ \hline
\end{tabular}}

\end{table}

\begin{figure}[t]
  \centering
   \includegraphics[width=.9\linewidth]{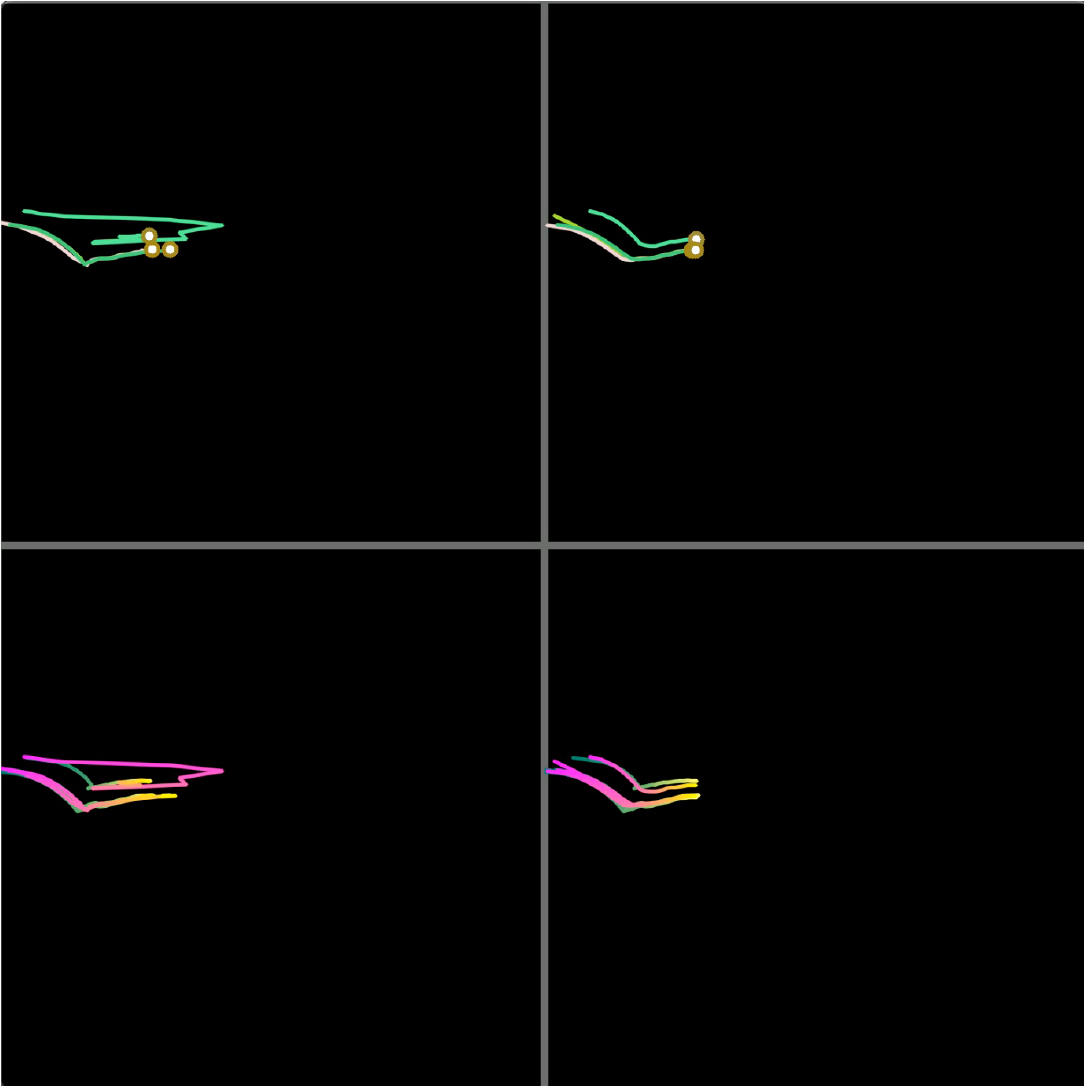} 

   \caption{Comparison between CoTracker and our method's predictions in the new view (i.e. 45\degree). \textbf{Upper-left}: CoTracker's results. \textbf{Upper-right}: Our results. \textbf{Lower-left}: Cotracker's \pinktext{predictions} overlapped with \greentext{GT} trajectories. \textbf{Lower-right}: our \pinktext{predictions} overlapped with \greentext{GT} trajectories. We show more results in  \textbf{our video}.}
   \label{fig: trajectory}
   % \vskip -0.15in
\end{figure}

\begin{figure}[t]
  \centering
   \includegraphics[width=.9\linewidth]{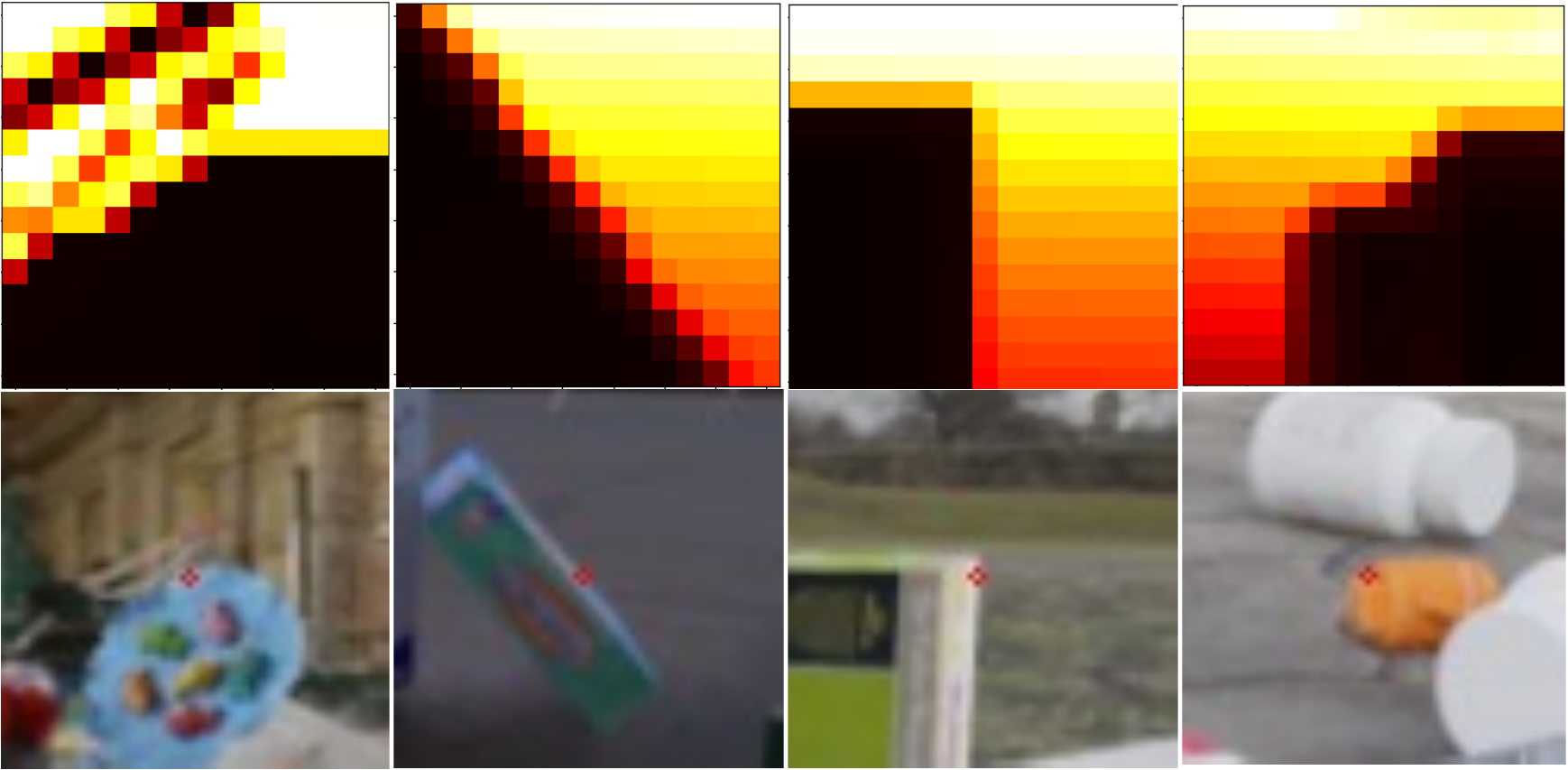} 

   \caption{Local patches of the GT depth maps used in the original view. These patches are cropped around points whose projections in the new view move abruptly.}
   \label{fig: heat_map}
   % \vskip -0.15in
\end{figure}

When analyzing the results of CoTracker in the new view - Fig.~\ref{fig: trajectory}, we observe many cases where the points move in zig-zag patterns or have  abnormally large motions at some specific time steps. We examine the depth values around these points in the original view and find that they typically reside on or near the object boundary - Fig.~\ref{fig: heat_map}. Consequently, if the predicted positions of CoTracker for the query points are slightly off by a few pixels, it can lead to a significant discrepancy in the extracted depth. This discrepancy causes the queries to oscillate when they are projected into different views.

% \FloatBarrier
\section{Ablation - Nearest and Bilinear Interpolation for Lifting into 3D}
\label{sec: nearest_interp}
In this section, we show the results of CoTracker~\cite{karaev2023cotracker} and TAPIR~\cite{doersch2023tapir} when we use the nearest neighbor to extract the depth value for each query point from the GT depth map - \textbf{Table~\ref{tab:BIvsNI}}. From these results, it can be shown that the nearest neighbor can extract the  depth value for the query more precisely and  improve the overall results of CoTracker and TAPIR consistently. However, despite the improvement, CoTracker and Tapir still underperform compared to ours in terms of the 3D metrics.

\begin{table}
    \centering
    \caption{Ablation on interpolation types for lifting 2D into 3D. BI: Bilinear Interpolation. NI: Nearest Neighbor}
    \label{tab:BIvsNI}
    \begin{tabular}{|c|c|c|c|}
    \hline
         & OA & 3D - $\delta^{avg}$ & $\delta^{avg}_{occluded}$ \\ \hline
        CoTracker (BI) & 92.5 & 57.8 & 8.7 \\ \hline
        CoTracker (NI) & 92.5 & 58.7 & 9.7 \\ \hline
        TAPIR (BI) & \textbf{96.5} & 46.9 & 3.8 \\ \hline
        TAPIR (NI) & \textbf{96.5} & 47.8 & 4.4 \\ \hline
        Ours & 93.4 & \textbf{73.1} & \textbf{44.0} \\ \hline
    \end{tabular}
    
\end{table}
% \FloatBarrier

\section{Ablation - 2D Loss Fine Tuning}
To compare with other 2D baselines, we have to project our 3D results into 2D.
To reduce the effect of the numerical error when projecting, we tried to tune our model directly in 2D with the projection loss:
\begin{equation}
    L^{2D}_{track} = \frac{1}{Tn_q}\sum_{l=1}^L \sum_{t=1}^T \sum_{i=1}^{n_q} \alpha^{l-1}  |q^{2D}_{t,i}-\hat{q}^{2D}_{t,i}|_1 \nonumber
\end{equation}
where $q^{2D}_{t,i}$ is the projection of the predicted 3D position of the query point, and $\hat{q}^{2D}_{t,i}$ is the 2D ground truth. Thanks to this, the accuracy in 2D is slightly improved. The ablations in the main paper were done without this projection loss hence the numbers were a little lower than tables 5 \& 6 with the 2D results.

\begin{table}
    \centering
    \caption{Ablation on the Projection Loss to reduce projection error.}
    \label{tab:my_label}
    \begin{tabular}{|c|c|c|}
        \hline
         & With Projection Loss & Without Projection Loss \\ \hline
        $\delta^{avg}$ & 87.8 & 87.0 \\ \hline
    \end{tabular}
    
\end{table}
\FloatBarrier

\end{document}